\documentclass{article}

\usepackage{arxiv}

\usepackage[utf8]{inputenc} 
\usepackage[T1]{fontenc}    
\usepackage{hyperref}       
\usepackage{url}            
\usepackage{booktabs}       
\usepackage{amsfonts}       
\usepackage{nicefrac}       
\usepackage{microtype}      
\usepackage{graphicx}

\usepackage{array}
\usepackage{amsmath}
\usepackage{amssymb}
\usepackage{latexsym}
\usepackage{subcaption}

\usepackage{cleveref}

\title{An Ontology Design Pattern for representing Recurrent Situations}

\author{Valentina Anita Carriero\\
Department of Computer Science and Engineering \\
University of Bologna \\
Mura Anteo Zamboni 7, 40126 Bologna, Italy \\ 
\texttt{valentina.carriero3@unibo.it} \And
Aldo Gangemi\\
Digital Humanities Advanced Research Centre \\
Department of Classical Philology and Italian Studies \\
University of Bologna \\
Via Zamboni 32, 40126 Bologna, Italy \\
\texttt{aldo.gangemi@unibo.it} \And
Andrea Giovanni Nuzzolese \\
Semantic Technologies Laboratory \\
Institute of Cognitive Sciences and Technologies \\
Italian National Research Council \\
Via San Martino della Battaglia 44, 00185 Rome, Italy \\ 
\texttt{andreagiovanni.nuzzolese@.cnr.it} \And 
Valentina Presutti \\
Department of Modern Languages, Literatures, and Culture \\
University of Bologna \\
Via Cartoleria 5, 40124 Bologna, Italy \\ 
\texttt{valentina.presutti@unibo.it}
}

\begin{document}
\maketitle

\begin{abstract}
In this paper, we present an Ontology Design Pattern for representing situations that recur at regular periods and share some invariant factors, which unify them conceptually: we refer to this set of recurring situations as \emph{recurrent situation series}. The proposed pattern appears to be foundational, since it can be generalised for modelling the top-level domain-independent concept of recurrence, which is strictly associated with invariance. The pattern reuses other foundational patterns such as Collection, Description and Situation, Classification, Sequence. Indeed, a recurrent situation series is formalised as both a collection of situations occurring regularly over time and unified according to some properties that are common to all the members, and a situation itself, which provides a relational context to its members that satisfy a reference description. Besides including some exemplifying instances of this pattern, we show how it has been implemented and specialised to model recurrent cultural events and ceremonies in ArCo, the Knowledge Graph of Italian cultural heritage. 
\end{abstract}


\section{Introduction}
\label{valentinacarriero:intro}
    In this paper, we investigate the concept of recurrence as applied to a series of situations, which \emph{recur}, i.e. they occur periodically or repeatedly, either by design (e.g. a festival), or by featuring similar attributes (e.g. a bird migration pattern); hence the name \emph{recurrent situations}. 
    Recurrence is associated with \emph{invariance}. We typically recognise something as a known entity when it is invariant under appropriate transformations (in time, space, context, etc.). This is a basic biological competence, typically starting in human babies at two months of age. 
    As we recognise an invariant object, we can also recognise an invariant situation, when multiple sets of related objects feature similar patterns.
    While this ability in humans develops into foundational concepts such as \emph{permanence}, \emph{similarity} and \emph{difference}, \emph{type}, etc., and is well investigated in cognitive science (see e.g. \cite{DBLP:journals/semweb/Gangemi20} for links between current cognitive science theories and knowledge representation), recurrence is less studied in ontology design. In particular, recurrence under time transformation is not fully addressed and captured by state-of-the-art patterns and ontologies.
    
    In this paper, we deal with \emph{constructed} and \emph{natural} recurrence of situations under time transformation. By \emph{constructed} we mean a recurrence created by social norms (e.g. elections or festivals) or emerging and established in local or even individual behavioral patterns (e.g. social meetings or daily routines). By \emph{natural} we mean a recurrence observed in nature (e.g. the raising of the sun, species migrations). By \emph{situation} we mean any eventuality or structure that someone intends to talk about (consider, refer to, mention), e.g. a conference, a system configuration, a trial.
    
    Existing ontologies mostly model recurrent situations as a special type of event, which represents its (repeatable) occurrences as generic parts of a more general event, neglecting their belonging to a \emph{series}, intended as a collection, whose members are temporally ordered. This is relevant and characterises the ontological nature of such a unifying entity.
    
    The Recurrent Situation Series Ontology Design Pattern (ODP), explained in the rest of this paper, addresses this issue. This pattern models collections of situations, in which the member situations: (i) are ordered in a temporal sequence, (ii) are separated by time intervals that obey some function (that is in principle computable), and (iii) share at least one invariance, unifying factor (e.g. a name, a topic, a purpose, etc.). Unifying factors make a recurrent situation series distinct from others. 
    This is a foundational pattern, since it models the top-level, domain-independent concept of recurrence, as applied to a series of situations.

The paper is organised as follows: Section \ref{valentinacarriero:rec-situations} introduces the problem addressed; Section \ref{valentinacarriero:abstract-odp} describes the pattern at an abstract level; Section \ref{valentinacarriero:implementations} presents two actual implementations of the pattern; Section \ref{valentinacarriero:usage} provides two usage examples. Finally, Section \ref{valentinacarriero:stateoftheart} reports the state of the art and Section \ref{valentinacarriero:conclusion} concludes the paper envisaging future work.

\section{Problem addressed: recurrent situations}
\label{valentinacarriero:rec-situations}
The \emph{Palio di Siena} is a horse race held twice each year on predefined dates 
in Siena, Italy. Preceded by other events, e.g. an historical costume parade, the race itself involves ten horses and riders and consists of three laps of \emph{Piazza del Campo}. The first horse that crosses the finish line, even without its rider, 
is declared the winner and awarded a banner of painted silk (\emph{palio}). Indeed, by riding bareback, more than one rider is usually thrown off his horse each year during the turns.

The Palio held on 2 July 2019 and the fall of the jockey \emph{Brigante} from his horse during that Palio are both \emph{situations}. Here, we refer to the notion of situation as a generalisation over any relational concept, e.g. events, states, actions. Both situations show a kind of repetition: indeed, every year the Sienese horse race occurs, and, most likely, the same can be said for the riders thrown off their horses.
Nevertheless, it is possible,  at  both  common-sense  and  philosophical level, associate only the first situation with a notion of recurrence that is strictly related to invariance: in recurrent situations, we can recognise a pattern in the iteration, at periodic intervals, of some properties that occur in each situation. While the Palio is explicitly designed, based on a plan, to iteratively occur at specific dates and following regulations (e.g. ten participants, crossbred horses), the falls of the jockeys during the horse race have a high probability, which can be empirically observed, to occur frequently, possibly at each race, but this observation is statistical in nature and is based on neither a plan nor a rule that regulates and assures its repetition with specific characteristics and at regular time intervals. Therefore, the annual race is designed as part of a \emph{series} with an inherent continuity, while the fall of a specific jockey is perceived as an episode.

There are two main distinctions that can be made between different types of recurrent situations, based on the origin of the regular recurrence: (i) natural vs social, (ii) unplanned vs planned. 

\noindent Natural recurrent situations are not constructed by social norms or practices: humans can observe them in very different ways, but their occurrence is not caused by the mere acceptance of social rules.
For example, morning sunrise, evening sunset, heartbeat, periodic migration of groups of animals from a region to another for feeding or breeding, changing seasons during the calendar year. We can be aware or not, we can provide alternative theories, but those things are not happening because we decree they should exist.

\noindent Social situations instead are man-made situations that have a specific purpose in a social setting, whose recurrence is set by humans: a world day, a train schedule, festivals, religious holidays, awards, sporting events, parliamentary elections, etc.

\noindent Planned situations are those defined by a clear plan, which regulates how the things involved in that plan shall be carried out; on the contrary, unplanned situations have not a predefined setting, and their properties are typically observed after some occurrences.

When we informally (in conversations) refer to a regular repetition of a situation (e.g. an edition of the Palio), the term \emph{recurrent event} is often used, as we are talking about an event with invariant properties that occurs repeatedly. The term \emph{event} however may cause confusion as it is attributed in literature several different and very specific meanings. The concept \emph{situation} generalises over them, addressing the large spectrum of possible entities that can recur, either for construction or by nature.  

Going back to our example, the \emph{Palio di Siena} is a series of consecutive situations that belong, as members, to a collection characterised by some unifying factors, i.e. properties that give those situations a unity. 

\section{The Recurrent Situation Series Ontology Design Pattern}
\label{valentinacarriero:abstract-odp}
In this section, we describe the abstract model of recurrent situation series, which is represented as an Ontology Design Pattern, according to OWL constructs.

\paragraph{Recurrent situation series.} A recurrent situation series is a set of situations repeating regularly over time. It is intended as a \emph{collection}, since it can be seen as a social object that depends on the collected items, i.e. its members \cite{Bottazzi2006}. 
A recurrent situation series is a particular type of collection where all the members are situations: entities flowing in time, either in the physical or social world, located in place, and involving some objects.
The members of a recurrent situation series, i.e. the recurring situations, must all play at least the role of `being a member of a recurrent situation series', and share one or more common properties.

\noindent At the same time, a recurrent situation series is a \emph{situation} itself, intended as a context in which a set of entities is  contextualized based on a \emph{description} or ``frame''. A recurrent situation series is similar to a plan, which defines how the things involved in that plan (i.e. the specific situations) shall be carried out, or follow some not completely predefined rules, e.g. where the situations are located, in which time of the year, etc.

\paragraph{Unifying criteria.}
The individual (re)occurring situations, while preserving their identity, are kept together and form a new entity, i.e. a recurrent situation series. In \cite{Bottazzi2006}, collections of items are conceptually unified by \emph{unity criteria}. A unity criterion is a concept, a social object created for collecting existing entities, representing a role that can be played by different entities. 
Therefore, a recurrent situation series is a collection, whose members are \emph{unified} by at least one unifying factor, i.e. organized according to one (or more if it evolves through time) recognizable pattern shared by all members. For example, unifying factors of Palio di Siena may be its name and the holding of a horse race in Piazza del Campo.
A set of situations can be perceived as a unitary collection when at least one element or property occurs in each situation member of the series, making all the situations ascribable to a homogeneous collection. For instance, all the situations member of a recurrent situation series can be placed in the same location, be about the same topic, be inspired by the same theme, establish certain activities, involve the same community, etc.
\noindent A recurrent situation series must have at least one unifying factor. Considering that, in theory, according to the Description and Situation ODP~\cite{DBLP:conf/coopis/GangemiM03}, every situation satisfies some description. We can expect that all members of a recurrent situation series satisfy the same description. Therefore, we can assert that it exists a description that plays the role of unifying factor for a recurrent situation series. This description is the one satisfied by all situation members of the series.  

A description is a socially-created object that provides a conceptualisation or a view on a situation. A recurrent situation description (RSD hereafter) either defines specific concepts, e.g. \texttt{repetition interval}, or (re)uses existing ones, e.g. \texttt{location}, which are supposed to \emph{classify} the entities involved in the recurrent situation, e.g. \texttt{6 months} for \texttt{repetition interval} or \texttt{Siena} for \texttt{location}. 

\noindent A RSD looks like a plan for recurrent situations: it unifies the recurrent situations series, describes its member situations, and is satisfied by these situations: this implies that at least some concepts in the RSD classify the entities (e.g. \texttt{6 months}, \texttt{Siena}) involved in the situations.

\noindent These unifying factors may be more or less flexible in unifying the recurrent situation series: properties that usually occur in all the situations member of the same collection can undergo change under particular conditions. For instance, yearly recurring situations may be usually held on a specific month of the year, but they still remain members of the same series even if one member situation takes place in another period of the year due e.g. to unusual climatic conditions. E.g., the FIFA world cup has always been in May/June/July since 1930, but in Qatar 2021 edition it will be held in December, since the summer is too hot in Qatar.
Moreover, a collection of situations recurring on an annual basis remains the same entity even if one specific edition is cancelled due to exceptional events.

\paragraph{Temporal sequence.}
As opposed to other types of collections, a partly predictable temporal interval holds between the members of a recurrent situation series. The members of the collection are all consecutive situations, thus creating a temporally-ordered sequence, so that each situation may have either previous situations, or next situations, or both, unless the recurrent situation series, planned to be a series of at least two situations, is interrupted after the first situation, or is never instantiated.

\noindent In the context of this pattern, the temporal interval is relevant only between members of the same recurrent situation series, while the sequence between a member situation and any other situation is out of the scope of this pattern.

\paragraph{Regular time period.} A collection of situations can be defined recurrent if its members occurrence are separated by regular time periods.
As for natural situations, this time period is observed in nature, while social situations usually define the recurrence in a plan, explicitly.
The \texttt{time period} concept works as a parameter that classifies typical sets of values: for example, a ``yearly'' time period classifies approximated time intervals, e.g. between 11 and 13 months instead of exactly 12.

\noindent Even considering approximation, there is a difference between the estimated time period, i.e. the time interval that is planned (for social and planned situation series) or foreseen (for natural and unplanned situation series) when the recurrent situation series starts, and the actual time period, i.e. the time interval that can be observed after the series has been instantiated and has multiple members. For instance, to a planned time period of ``approximately 2 years'' corresponds a measured time interval that represents the average of time intervals actually occurring between the situations, and can vary significantly if e.g. some yearly situations are postponed or canceled. This measurement could be especially relevant to recurrent situations series, when the time period is a requirement to satisfy (e.g. a periodical journal), or when a time period different from the expected is due to negative external causes (e.g. natural climate changes).

\section{Implementations of the Recurrent Situation Series ODP}
\label{valentinacarriero:implementations}
The abstract pattern, as described in Section \ref{valentinacarriero:abstract-odp}, can be formalised and possibly implemented as an Ontology Design Pattern (ODP), in order to be reused in different contexts. In this section, we report as an example two implementations of this pattern: the ODP that has been published on the \emph{ODP portal}\footnote{http://ontologydesignpatterns.org} \cite{Carriero2019WOP}, and an implementation specific to the Cultural Heritage (CH) domain, in the context of the  ArCo\footnote{https://w3id.org/arco/ontology/arco} ontology network \cite{Carriero2019}.

\subsection{ODP catalogue}
The Recurrent Situation Series pattern\footnote{\texttt{rss:} http://www.ontologydesignpatterns.org/cp/owl/recurrentsituationseries.owl\#} reuses five existing Ontology Design Patterns (ODPs): \emph{Collection}\footnote{http://ontologydesignpatterns.org/wiki/Submissions:Collection}, \emph{Situation}\footnote{http://ontologydesignpatterns.org/wiki/Submissions:Situation},
\emph{Description and Situation}\footnote{http://www.ontologydesignpatterns.org/cp/owl/descriptionandsituation.owl},
\emph{Classification}\footnote{http://ontologydesignpatterns.org/wiki/Submissions:Classification}, \emph{Sequence}\footnote{http://ontologydesignpatterns.org/wiki/Submissions:Sequence}. These patterns have been extracted from DOL\-CE+DnS Ultra-lite\footnote{\texttt{dul:} http://www.ontologydesignpatterns.org/ont/dul/DUL.owl\#} foundational ontology, a lighter version of DOLCE+DnS~\cite{Gangemi2002}, thus our ODP is also aligned to it. Moreover, all reused ODPs are annotated with the OPLa ontology~\cite{DBLP:conf/semweb/HitzlerGJKP17}, which facilitates future reuse of this pattern by supporting the process to understand and explore the ODP. For instance, OPLa allows us to express that a pattern in a local ontology is a specialisation or a generalisation of another pattern.

\begin{table}[t]
	\caption{Competency questions answered by the Recurrent Situation Series ODP.}
	\label{tab:recurrent-situation-cq}
	\begin{tabular*}{0.9\textwidth}{ll}
		\hline
		{ ID } & { Competency question }\\\hline 
		CQ1 & Which are the situations of a recurrent situation series?\\
		CQ2 & Which is the time period elapsing between two situations of a recurrent situation series? \\
		CQ3 & When is the next situation of a recurrent situation series scheduled? \\
		CQ4 & What are the unifying criteria of a recurrent situation series? \\
		CQ5 & Which is the temporal validity of a unifying factor of a recurrent situation series? \\
		CQ6 & Which is the description satisfied by all the situations of a recurrent situation series? \\
		CQ7 & Which is the (immediate) next situation in a recurrent situation series? \\
		CQ8 & Which is the (immediate) previous situation in a recurrent situation series? \\\hline
	\end{tabular*}
\end{table}

\begin{figure}[ht!] 
\centering
\includegraphics[width=\textwidth]{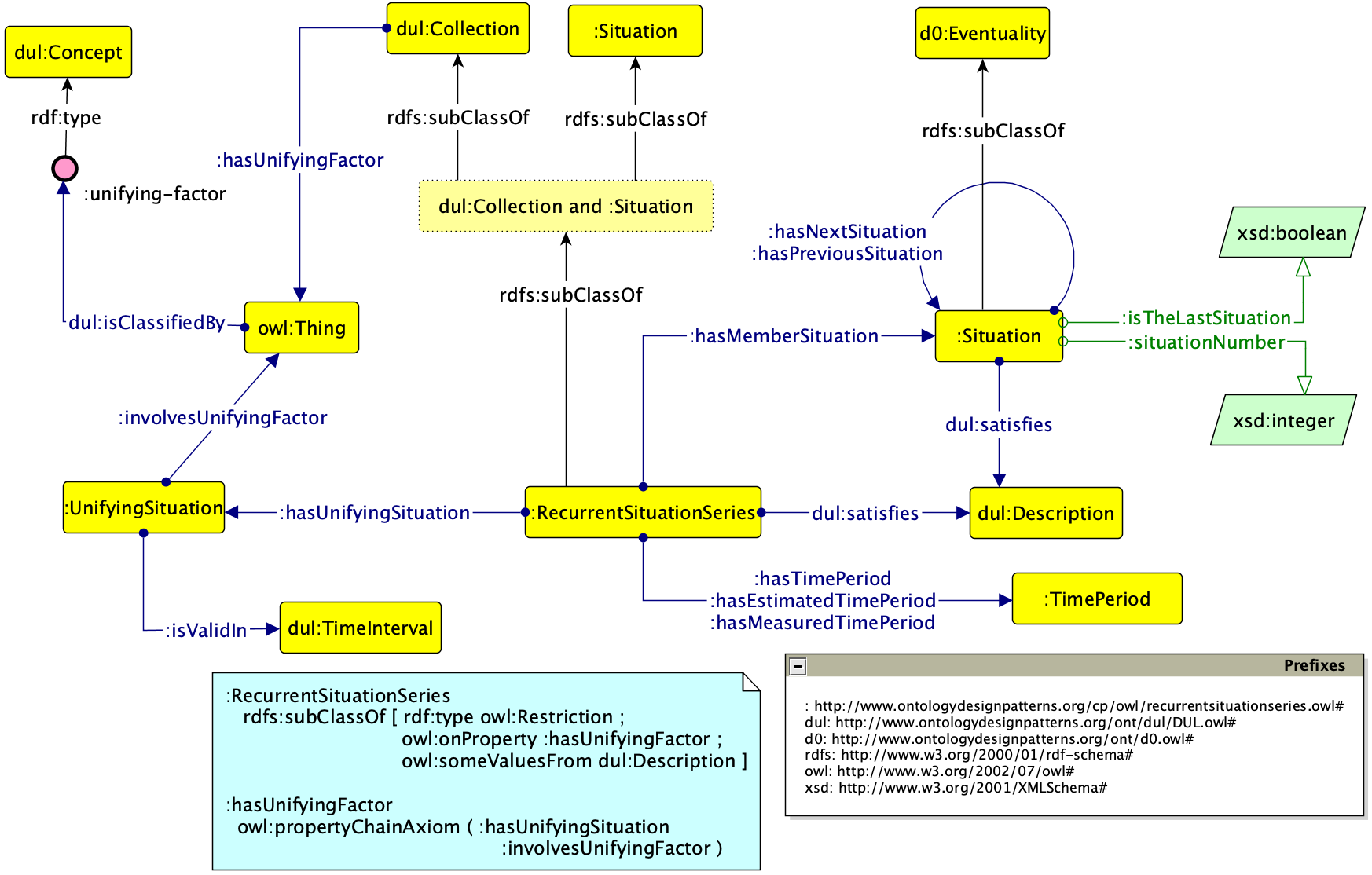}
	\caption{Pattern for Recurrent Situation Series, in Graffoo notation (https://essepuntato.it/graffoo/\allowbreak{}specification/).
	A recurrent situation series is related to: its member situations, which are temporally ordered in sequence; one or more unifying factors; one or more unifying situations, which involve unifying factors and their temporal validity; the estimated and measured time period between two consecutive situations. At least one unifying factor of the series is a description that is satisfied by each situation member of the series. Moreover, the series satisfies a description defining the concepts characterising recurrent situation series.}
	\label{img:recurrent-situation}
\end{figure}

The ODP, as submitted to the ODP portal, can be found at \texttt{http://ontology\allowbreak{}design\allowbreak{}patterns.\allowbreak{}org/\allowbreak{}wiki/\allowbreak{}Submissions:\allowbreak{}Recurrent\allowbreak{}Situation\allowbreak{}Series}.
A diagram depicting the pattern is shown in Figure \ref{img:recurrent-situation}, and the Competency Questions (CQs) this pattern can answer are reported in Table ~\ref{tab:recurrent-situation-cq}.
The general pattern provides a reusable template for modelling recurrent situation series. For the concepts of situation and time period, which are not native to this pattern, it reuses the semantics associated with: \texttt{d0:Eventuality}, from the foundational ontology DOLCE-Zero\footnote{\texttt{d0:} http://www.ontologydesignpatterns.org/ont/d0.owl\#}, as a top-level concept for any situation, event or activity, and the formalisation of time periods as defined by the \emph{Time Period} ODP\footnote{http://ontologydesignpatterns.org/wiki/Submissions:TimePeriod}.
In the following, we describe classes and properties of the pattern.

\paragraph{RecurrentSituationSeries.} Recurrent situations are represented as individuals of the class \texttt{:RecurrentSituationSeries}, which is modeled as the intersection of the classes \texttt{dul:Collection} and \texttt{:Situation} (axiom 1): a recurrent situation series is both a collection of situations, and a situation on its turn, extending along the time in which its members occur (see below).

\begin{eqnarray}
\texttt{RecurrentSituationSeries}  & \sqsubseteq & \texttt{dul:Collection} \sqcap \texttt{Situation}
\end{eqnarray}

\paragraph{hasMemberSituation.} The membership for a collection of recurring situations (a \texttt{:Recurrent\-Situation\-Series}) is restricted to situations only (axiom 2), and a specific object property \texttt{:has\-Member\-Situation} is provided as a pattern vocabulary item. Its inverse property is \texttt{:is\-Situation\-Member\-Of}. A recurrent situation series may also have 0 members, in order to envision series that are never instantiated (axiom 3). As aforementioned, the concept of situation is not native to this pattern, thus we introduce a generic class \texttt{:Situation} as a subclass of \texttt{d0:Eventuality} (axiom 4).

\begin{eqnarray}
\texttt{RecurrentSituationSeries} & \sqsubseteq & \forall \texttt{hasMemberSituation.Situation}\\
\texttt{RecurrentSituationSeries} & \sqsubseteq & \geqslant0~ \texttt{hasMemberSituation.Situation}\\
\texttt{Situation} & \sqsubseteq & \texttt{d0:Eventuality}
\end{eqnarray}

\paragraph{UnifyingFactor.} \texttt{:UnifyingFactor} is modelled as a subclass of \texttt{dul:Concept}, since it is a social object for collecting situations. The \texttt{:Recurrent\allowbreak{}Situation\allowbreak{}Series} is related to the invariant unifying factors by means of the object property \texttt{:has\allowbreak{}Unifying\allowbreak{}Factor}. An existential axiom (axiom 5) states that it exists a \texttt{dul:\allowbreak{}Description} playing as unifying factor for a recurrent situation series. This description defines concepts (\texttt{dul:\allowbreak{}defines}) that conceptualise all the situations that are member of the series, thus it \texttt{dul:is\allowbreak{}Satisfied\allowbreak{}By} them. Therefore, it works as a unifying criterion for the whole series. Moreover, the recurrent situation series itself satisfies another description, which defines all the concepts conceptualising a recurrent situation series (e.g. the general concept of unifying factor, the concept of time period, etc.).

\begin{eqnarray}
\texttt{RecurrentSituationSeries} & \sqsubseteq & \exists \texttt{hasUnifyingFactor.dul:Description}
\end{eqnarray}

\paragraph{UnifyingSituation, isValidIn.} The validity of a unifying factor can be limited to a specific time interval, for criteria that unify the series for a limited period of time, due to their flexibility or external causes. For modelling the temporal validity of a unifying factor, we need a temporally-indexed situation (\texttt{:\allowbreak{}Unifying\allowbreak{}Situation}) involving a unifying factor (axiom 6). The relation between this unifying situation and the time interval of its validity is expressed through the \texttt{:isValidIn} object property (axiom 7).

\begin{eqnarray}
\texttt{UnifyingSituation} & \sqsubseteq & \exists \texttt{involvesUnifyingFactor.owl:Thing}\\
\texttt{UnifyingSituation} & \sqsubseteq & \exists \texttt{isValidIn.dul:TimeInterval}
\end{eqnarray}

\paragraph{hasNextSituation, hasPreviousSituation.} It is possible to represent which is the previous and the next situation of a particular member of the series by means of the object properties \texttt{:hasPreviousSituation} and \texttt{:hasNextSituation}. The two properties are further specialised into \texttt{:hasImmediatePrevious\allowbreak{}Situation} and \texttt{:has\allowbreak{}Immediate\allowbreak{}Next\allowbreak{}Situation}, respectively, in order to associate the situations with their immediate next and previous situations in the sequence.

\noindent These properties express a sequential relation between situations that are members of the same recurrent situation series, while they do not relate a situation to any other previous or next situation. This intended restriction cannot be expressed in OWL 2 DL~\cite{Hoekstra2008}, because this requires a ``diamond'' construction with explicit coreferent variables. However, it is possible to check the local integrity of an instantiation of a \texttt{:RecurrentSituationSeries} by using inference rules.

A situation (member of a recurrent situation series) can be linked by the property \texttt{:hasNextSituation} to another situation, which is also member of a recurrent situation series, if both situations are member of the same collection. For instance, let us consider the following SPARQL query. A triple asserting a local inconsistency between the two recurrent situation series will be generated, if they are not the same or if there is not an identity relation (\texttt{owl:sameAs}) between the two recurrent situation series. Similar SPARQL queries can be executed for the related constraints (e.g. situations connected by \texttt{:hasPreviousSituation} object property). These queries can be embedded in the RDF model via SPIN\footnote{https://spinrdf.org/} or SHACL\footnote{https://www.w3.org/TR/shacl/}.

\begin{verbatim}
 PREFIX rss: <http://www.ontologydesignpatterns.org/cp/owl/
                                    recurrentsituationseries.owl#>
 CONSTRUCT {?re1 rss:isLocallyInconsistentWith ?re2}
 WHERE {
        ?rss1 rss:hasMemberSituation ?sit1 . 
        ?rss2 rss:hasMemberSituation ?sit2 . 
        ?sit1 rss:hasNextSituation ?sit2 . 
        filter not exists {?rss1 owl:sameAs+ ?rss2}
        filter (?rss1 != ?rss2)
        filter (?sit1 != ?sit2)
       }
\end{verbatim}

The datatype property \texttt{:isTheLastSituation}, with range \texttt{xsd:boolean}, expresses whether a situation is the last of the series or not, while \texttt{:situationNumber} represents the number of the situation within the sequence (e.g. ``2'' for the 2nd situation).

\paragraph{TimePeriod.} While the time period is a core element of this pattern, it can be modelled   in different ways, depending on the local requirements. Thus, we introduce a generic class \texttt{:TimePeriod}. Nevertheless, we propose a possible solution for implementing the pattern, by reusing the class \texttt{tp:TimePeriod} from the \emph{TimePeriod} ODP\footnote{\texttt{tp:} http://www.ontologydesignpatterns.org/cp/owl/timeperiod.owl\#} in our usage examples (see Section \ref{valentinacarriero:usage}). A \texttt{tp:TimePeriod} is related to a measurement unit and a measurement value. The collection of situations is related to the recurrent time period with the object property \texttt{:hasTimePeriod} (axiom 8), which is defined as a property chain \texttt{[:hasMemberSituation} $\circ$ \texttt{:hasTimePeriodBeforeNextSituation]}. The object property \texttt{:hasTimePeriodBeforeNextSituation} relates a situation, member of a recurrent situation series, to the approximate (e.g. yearly, monthly, etc.) time period that typically elapses before the next situation, and is different from the actual time interval between two specific situations of the series, which can be derived from the time intervals measured between any two members. \texttt{:hasTimePeriod} is indeed further specialised into \texttt{:hasEstimatedTimePeriod} and \texttt{:hasMeasuredTimePeriod}.

\begin{eqnarray}
\texttt{RecurrentSituationSeries} & \sqsubseteq & \exists \texttt{hasTimePeriod.TimePeriod}
\end{eqnarray}

\paragraph{TimePeriodMeasurementUnit.} The \texttt{tp:TimePeriod} is related to the unit for measuring it, \texttt{tp:TimePeriodMeasurementUnit} (e.g. year, month, week), by means of the object property \texttt{tp:hasTimePeriodMeasurement\allowbreak{}Unit}, and to the time value through the datatype property \texttt{tp:timePeriodValue}, with range \texttt{xsd:integer}. 


\subsection{ArCo: the Italian cultural heritage Knowledge Graph}
A second implementation of the pattern is given in ArCo (Architecture of Knowledge)\footnote{https://w3id.org/arco}, the Italian Cultural Heritage (CH) knowledge graph~\cite{Carriero2019}. We have been working on formally representing recurrent situations while developing this KG, which consists of an ontology network and facts on Italian cultural properties, based on the General Catalogue (GC) maintained by the Italian Institute of the General Catalogue and Documentation (ICCD-MiBAC).
ArCo adopts an agile and iterative, pattern-based and test-driven ontology development methodology, called eXtreme Design (XD) \cite{Blomqvist2016}. Its requirements are provided in the form of user stories, as scenarios and real use cases, by a growing community of customers, including ICCD. 
In particular, two modelling issues led us to implement a solution for representing situations recurring regularly over time.

In the GC catalogue records, which describe Italian cultural properties, it is possible to find information about events (title, place, organizer) involving cultural properties, such as exhibitions. While there is no explicit information about possible recurrent cultural events, by analyzing the data, it is clear that there are many cases of collections of recurrent events (e.g. ``third exhibition", ``tenth painting award").

The second use case concerns a particular type of intangible cultural heritage: ceremonies, customs and celebrations related to the \emph{year cycle} (e.g. Carnival, Ramadan) or to the \emph{season cycle} (e.g. popular belief, science, myth, phenomena related to specific periods of the year), which recur regularly and whose periodicity is explicitly referred to in the GC catalogue records (e.g. annual, every two years, three times a year).

The Recurrent Situation Series ODP has been indirectly reused (i.e. has been reused as a template) in the Cultural Event\footnote{\texttt{a-ce:} https://w3id.org/arco/ontology/cultural-event} ontology module. It has been specialised in order to represent collections of situations that are cultural events (\texttt{cis:CulturalEvent}\footnote{\texttt{cis:} http://dati.beniculturali.it/cis/}) and exhibitions (\texttt{a-ce:Exhibition}).

\section{Use cases}
\label{valentinacarriero:usage}
In this section we introduce two use cases, and show how they can be addressed by the Recurrent Situation Series ODP. The first example is a  workshop, planned to occur periodically; the second example concerns a collection of recurrent situations observed in nature: annual animal migrations.

\subsection{Artificial recurrent situation series: the Workshop on Ontology Design and Patterns}

\noindent The Workshop on Ontology Design and Patterns (WOP) is a series of workshops that are co-located with ISWC (International Semantic Web Conference). ISWC started in 2002, WOP in 2009. Both the conference and the workshop recur yearly, between October and November, hence there is a time period of about one year between two situations of the same series. Nevertheless, it can happen that one edition is skipped: indeed, WOP has not been held in 2011. The location of these events may change every year. Usually, 3 geographical regions are involved: Europe, America, Asia/Oceania are sequentially alternated. However, unforeseen conditions can alter this pattern: in 2020, due to COVID-19 pandemic, they are organised as virtual events. 
The Workshop on Ontology Design and Patterns can be seen as a unitary collection on account of many (more or less) invariant factors: it is co-located with ISWC, its main topic is modular and pattern-based design, it involves a community mostly formed by computer scientists and ontologists, it is organised by the Ontology Design and Patterns Association (ODPA), etc.
Some factors are common to a subset of the situations member of the series, e.g. the name of the workshop changed 3 times: (i) \emph{Workshop on Ontology Patterns} from the first edition to 2012, (ii) \emph{Workshop on Ontology and Semantic Patterns} from 2013 to 2016, and (iii) \emph{Workshop on Ontology Design and Patterns} since 2017.
Moreover, the workshop recurs regularly: its estimated time period is of about 1 year, but the measured time period slightly differs, since it was not held in 2011, and its dates are not predefined.

\begin{figure}[ht!] \centering
\begin{subfigure}{\textwidth}
  \centering
  \includegraphics[width=\textwidth]{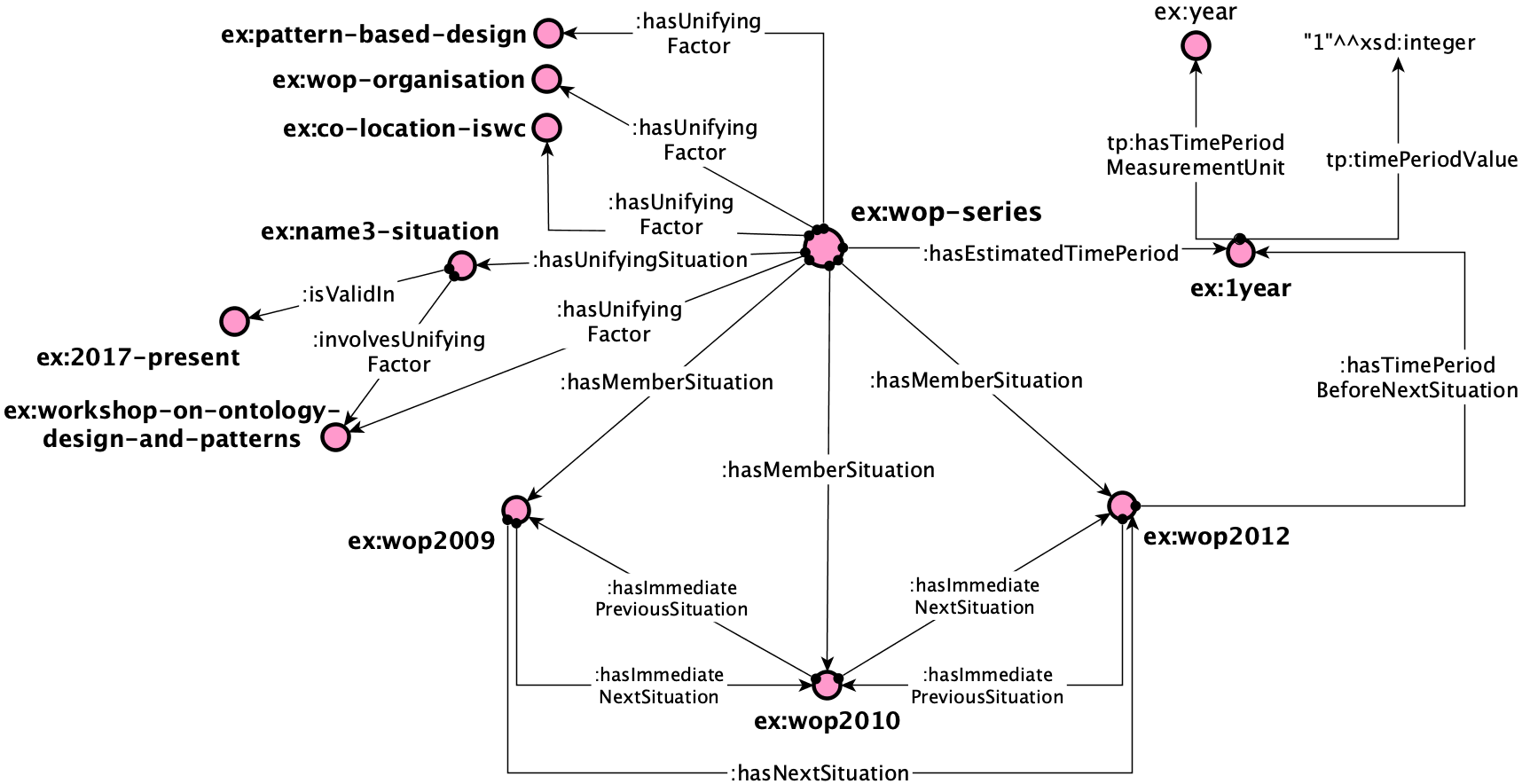}  
  \caption{The Workshop on Ontology Design and Patterns (WOP) is a recurrent situation series and is related to its first three consecutive situations; three unifying factors and one unifying situation with a limited temporal validity; the estimated time period between two consecutive events.}
  \label{fig:wop1}
\end{subfigure}
\begin{subfigure}{\textwidth}
  \centering
  \includegraphics[scale=0.40]{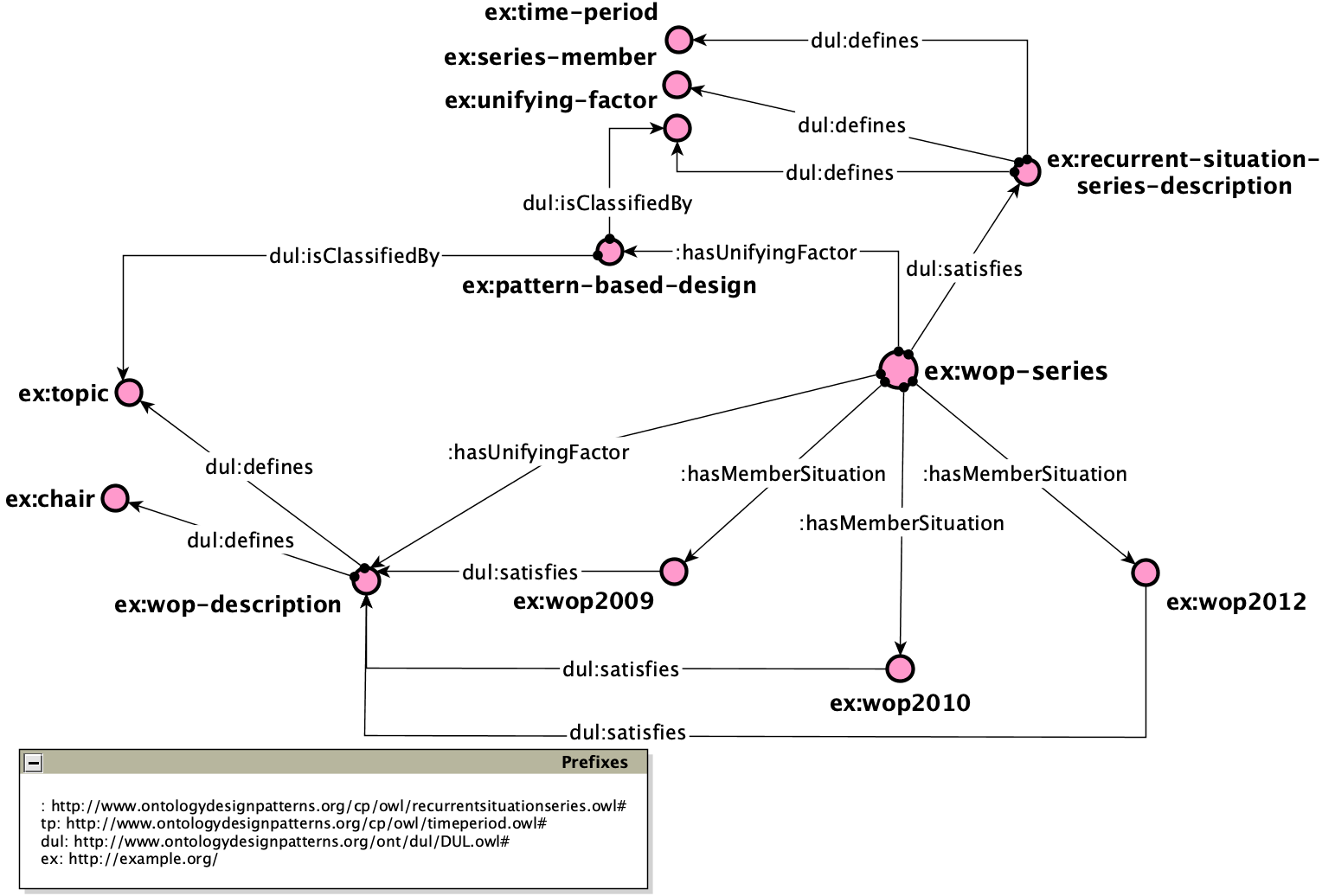}  
  \caption{The situations member of the WOP series satisfy a description; this description defines the concepts (topic, chair) characterising each workshop, and provides a unifying factor to the series. The series, in turn, satisfies a description defining the core concepts (time period, unifying factor, series member) of recurrent situation series.}
  \label{fig:wop2}
\end{subfigure}
\caption{The Workshop on Ontology Design and Patterns (WOP) is a recurrent situation series.}
\label{img:wop}
\end{figure}

Figure \ref{img:wop} depicts how WOP can be modelled by reusing this ODP. In Subfigure \ref{fig:wop1}, the series collecting the annual workshops is an instance of \texttt{:Recurrent\allowbreak{}Situation\allowbreak{}Series}, and is related to its members (e.g. WOP 2009, WOP 2010 and WOP 2012). The sequential relations between these (immediate) consecutive situations are modelled by \texttt{hasNextSituation} and \texttt{hasImmediateNextSituation}, and their inverses.
Since the beginning of the series, three unifying factors have been valid: the topic of the workshops (\texttt{ex:\allowbreak{}pattern-\allowbreak{}based-\allowbreak{}design}), the organising committee (\texttt{ex:wop-\allowbreak{}organisation}), and being held as joint event of ISWC (\texttt{ex:\allowbreak{}co-\allowbreak{}location-\allowbreak{}iswc}). So far, these unifying factors  do not necessarily need the specification of a temporal validity. Differently, the name of the workshops can be considered a unifying factor, however it has changed over time. This can be modelled by introducing a \texttt{:Unifying\allowbreak{}Situation} to specify the time interval in which e.g. the current name of the series is valid (\texttt{ex:2017-present}). 
The series is related to its estimated time period of \texttt{ex:1year}.

Subfigure \ref{fig:wop2} depicts the description satisfied by the members of the \texttt{ex:wop-series}, i.e. \texttt{ex:wop-description}. This description defines a number of concepts (the topic, the role of chair) that can be used for classifying the entities involved in the member situations (each workshop edition). For instance, the concept \texttt{ex:topic} classifies the entity \texttt{ex:pattern-\allowbreak{}based-\allowbreak{}design}, which is the main topic of the workshop(s). Indeed, the factors unifying the whole series may be classified by some or all the concepts defined in this description.
Finally, the WOP series satisfies a description that defines the core concepts of recurrent situation series (\texttt{ex:recurrent-\allowbreak{}situation-\allowbreak{}series-\allowbreak{}description}), such as the time period, the role of being a situation member of a series, the unifying factor.

\subsection{Natural recurrent situation series: the migrations of the Arctic terns}
Animal migration is the regular long-distance movement of animals, on an annual or seasonal basis, that may be caused by local climate, the season of the year, and food availability.
The Arctic tern (\emph{Sterna paradisaea}) is a long-distance migrant seabird: it is famous for periodically flying from the northern hemisphere, where it breeds, to the Southern Ocean and back again each year. In this way, it lives two summers a year. It departs the breeding site (Arctic) around August and starts its flight back from the Antarctic around March.
The migration of the Arctic tern is a unitary collection of situations since its member situations involve all the members of a specific bird species, happen along the North-South axis, depend on the season of the year in a specific place, etc.

\begin{figure}[ht!] \centering
\includegraphics[width=\textwidth]{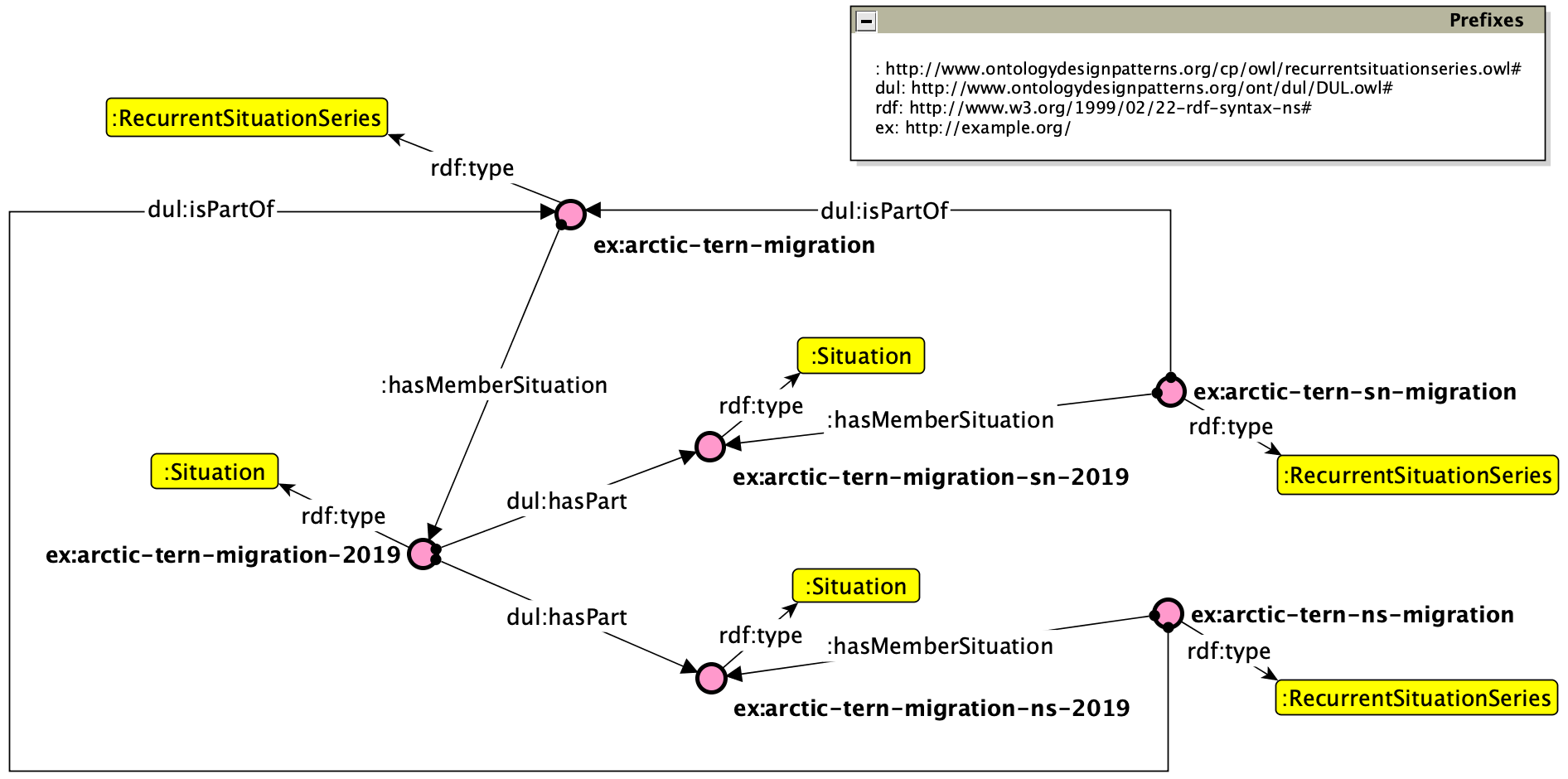}
	\caption{The Arctic tern migration is a recurrent situation series, with the annual migrations of the Arctic terns as members. Each annual migration consists in two parts, the migration from North to South and the migration from South to North, which are situations members of two separate recurrent situation series. Finally, these recurrent situation series are part of the annual Arctic tern migration.}
	\label{img:arctic-tern}
\end{figure}

Figure \ref{img:arctic-tern}, depicts how the migration of the Arctic ster is modelled with this ODP. The \texttt{ex:arctic-tern-migration} is an instance of \texttt{:Re\allowbreak{}cur\allowbreak{}rent\allowbreak{}Sit\allowbreak{}uation\allowbreak{}Series}, and represents the annual migration of the seabird. The series includes as members all the annual migrations, e.g. the \texttt{ex:\allowbreak{}arctic\allowbreak{}-tern-\allowbreak{}migration-\allowbreak{}2019} is the member situation representing the migration from North to South and back again happened in 2019. This situation actually consists in two parts (\texttt{dul:\allowbreak{}hasPart}), which are also situations: the flight from North to South in August 2019 (\texttt{ex:\allowbreak{}arctic-\allowbreak{}tern-\allowbreak{}migration-\allowbreak{}ns-\allowbreak{}2019}) and the flight from South to North in March 2019 (\texttt{ex:\allowbreak{}arctic-\allowbreak{}tern-\allowbreak{}migration-\allowbreak{}sn-\allowbreak{}2019}). These situations are in turn members of two separate recurrent situation series, also with an annual time period: the collection of all the flights from North to South and the collection of the ones from South to North. The \texttt{ex:arctic-tern-ns-migration} and the \texttt{ex:arctic-tern-sn-migration} are series part of the annual arctic tern migration series.

\section{Related work}
\label{valentinacarriero:stateoftheart}
Most work about recurrence in situations focuses on different types of recurrent situations and of constraints that bound a recurrent situation.
\cite{Ladkin1986} and \cite{Cukierman1996} investigate calendar-based periodic situations, e.g. repeating time intervals such as ``the first Friday each month''.
\cite{TuzhilinC95} distinguishes between strongly periodic events, intermittent events and nearly periodic events, with respect to the strength and the regularity of intervals between different occurrences.
\cite{Chakravarty99} defines a periodic pattern as the repetition of a component (e.g. an event) over time, such as daily glucose measurements. Both local (referring to a single component, e.g. the duration of one event of the pattern) and periodic (referring to the set of repeating events, e.g. the duration of the whole pattern) constraints can be applied to periodic patterns, which can be also split into sub-patterns applied only to a subset of the repeating events. Constraints related to single members of the periodic series and to the series as a whole are also considered in \cite{Terenziani02}.

The Recurrent Situation Series pattern follows the approach adopted by~\cite{Terenziani02}, where repeated events are considered as collections of events of the same type, and can be infinite in case no duration of the collection has been defined. Here, the notion of recurrent situation is related to the idea of a coherent series of events that become part of a pattern by means of their iteration, at regular but not necessarily equal time intervals.

The concept of collection is studied in \cite{Bottazzi2006}, but the focus of the authors is on collections of \emph{endurants}, i.e. entities that are directly localised in space such as objects and substances, while they do not deal with the properties peculiar to a collection of situations, as entities directly localised in time (\emph{perdurants}).

There exist some examples of ontologies modelling recurrent situations, but in most cases they are modeled as a particular type of event, rather than as a collection of situations or events.

In DBpedia\footnote{https://dbpedia.org/}, recurrent situation series such as annual film award ceremonies are represented as events, and DBpedia ontology\footnote{\texttt{dbo:} http://dbpedia.org/ontology/} models concepts such as \texttt{dbo:Olympics}, \texttt{dbo:Tournament} and \texttt{dbo:MusicFestival} as subclasses of \texttt{dbo:Event}, with an overlap between the series and its different editions.
For instance, the Palio di Siena\footnote{http://dbpedia.org/resource/Palio\_di\_Siena} is an instance of \texttt{dbo:SportsEvent}, and there are no links to specific resources representing its periodic editions.
DBpedia ontology also derives classes from YAGO ontology, which in turn are automatically derived from Wikipedia and WordNet, such as \texttt{yago:WikicatRecurringEventsEstablishedIn1895}\footnote{\texttt{yago:} http://dbpedia.org/class/yago/}, where the semantics of recurrence is only implied by the name of the class. DBpedia instances of recurrent situations are usually related to their frequency (e.g. “biennial, every two years”) through the \texttt{dbp:frequency}\footnote{\texttt{dbp:} http://dbpedia.org/property/} datatype property.

Wikidata defines the concept of recurring event (\texttt{wd:Q15275719}\footnote{\texttt{wd:} http://www.wikidata.org/entity/}), ``event recurring at an interval’’, as a specific type of event with an associated time period (\texttt{wdt:P2257}\footnote{\texttt{wdt:} http://www.wikidata.org/prop/direct/}) e.g. 4 years for the Olympic Games. \texttt{wd:Q15275719} is further specialised based on its frequency (e.g. biennial event) or its category (e.g. recurring sporting events). Wikidata relates different editions to their recurring event through the object property \emph{part of} (\texttt{wdt:P361}). Being this property transitive, a sub-event that is part of a specific edition (e.g. a talk by an author of a scientific paper that is part of an edition of an annual conference), would also be considered part of the annual conference as a recurring situation. 

A mereological approach is less clear in distinguishing typical features of recurrence, e.g. a recurrent situation may have no members at all and still be a recurrent situation, but, in this case, a mereological whole would be empty, i.e. without parts, which is both formally and intuitively impossible. 
Moreover, the intermittent presence of the situation would make it a part of a maximal continuous situation, which is not the same as a collection of specifically identified situations. In other words, if we admit that recurrence is a temporal mereological concept, we need also to admit maximal mereological wholes for any set of situations that have some temporally ordered similarity, which in the considered use cases would be an overcommitment.

As part of the \emph{pending section}\footnote{Section with ontology terms still work-in-progress, subject to possibly significant change: https://schema.org/docs/pending.home.html} schema.org defines the class \texttt{schema:EventSeries}\footnote{\texttt{schema:} http://schema.org/}: ``a collection of events that share some unifying characteristic'', modelled as a subclass of \texttt{schema:Event} and \texttt{schema:Series}. Included events are linked with the series using the \texttt{schema:superEvent} property, which expresses a part-whole relation, thus, as in Wikidata, both the series and the related events are modelled as events, being the latter part of the former. Even if the comment of the class cites the theme, location and organisers as typical examples of unifying characteristics, these are not modelled in the ontology.
\texttt{schema:eventSchedule} associates a series of repeating events with a \texttt{schema:Schedule}, which is the core class for defining the recurrence of an event series, i.e. its repeating time period. \texttt{schema:repeatFrequency} defines the frequency (\texttt{schema:Duration}) at which the events will occur according to a schedule, while \texttt{schema:repeatCount} represents the number of times a recurring event will take place. 

\section{Conclusion}
\label{valentinacarriero:conclusion}
In this paper, we explained an abstract pattern to represent situations that recur regularly over time, in a temporal sequence, and are members of a unitary collection since they share some invariant properties. This pattern can be generalised in order to represent the more general concept of recurrence. We also discuss two implementations of this pattern as an Ontology Design Pattern (ODP), a small ontology to be reused in different contexts. This pattern has been created for addressing the requirement of modeling cultural events, exhibitions, traditional ceremonies, festivals, etc. However, it seems to be useful for representing all recurring situations, which can be distinguished depending on the origin of their regular intermittence: (i) not artificial situations whose regular recurrence can be observed in nature by humans, but has a \emph{natural origin} (e.g. the sunrise every morning or periodic migrations of animals); (ii) man-made situations where the regular repetition and the unifying factors are \emph{set by humans} (e.g. a world day or a train schedule); (iii) situations created by humans with a periodicity regulated by a clear purpose, as part of a \emph{workflow} (e.g. a medical prescription defining time intervals to take a medicine or periodical supply of services).

In our future work, we plan to study all types of recurrent situations, to thoroughly investigate the general concept of recurrence, identifying possible new features to be represented, and to experiment and evaluate other possible patterns for recurrent situations, alternative to the one presented in this paper.

\bibliographystyle{ios1}   
\bibliography{reference.bib}
\end{document}